%% file: main.tex
\documentclass[10pt,twocolumn,letterpaper]{article}
\usepackage{graphicx}
\usepackage{epstopdf} 
\usepackage{subcaption}
\usepackage{float}
\usepackage{ulem}
\DeclareGraphicsExtensions{.png,.pdf,.jpg}
\usepackage[T1]{fontenc}
\usepackage[pagenumbers]{cvpr} 
\usepackage{fancyhdr}
\input{preamble}

\definecolor{cvprblue}{rgb}{0.21,0.49,0.74}
\usepackage[pagebackref,breaklinks,colorlinks,citecolor=cvprblue]{hyperref}
\usepackage{lipsum} 

\def\confName{CVPR}
\def\confYear{2024}

\title{Technical Approach for the EMI Challenge in the 8th Affective Behavior Analysis in-the-Wild Competition}

\author{Jun Yu\textsuperscript{1}, 
Lingsi Zhu\textsuperscript{1}, 
Yanjun Chi\textsuperscript{1}, 
Yunxiang Zhang\textsuperscript{1}, 
Yang Zheng\textsuperscript{1}, 
Yongqi Wang\textsuperscript{1}, 
Xilong Lu\textsuperscript{1}\\
\textsuperscript{1} University of Science and Technology of China\\
{\tt\small \{harryjun\}@ustc.edu.cn}\\
{\tt\small \{ls-zhu24 , yjChi , mesa , zhengyang , wangyongqi , luxilong\}@mail.ustc.edu.cn}
}

\begin{document}
\maketitle

\input{sec/0_abstract}    
\input{sec/1_intro}

\input{sec/2_related_work}
\input{sec/3_method}

\input{sec/4_experiment}

\input{sec/5_conclusion}

{
    \small
    \bibliographystyle{ieeenat_fullname}
    \bibliography{main}
}


\end{document}

%% file: preamble.tex
\usepackage[dvipsnames]{xcolor}


%% file: sec/0_abstract.tex
\begin{abstract}
Emotional Mimicry Intensity (EMI) estimation plays a pivotal role in understanding human social behavior and advancing human-computer interaction. The core challenges lie in dynamic correlation modeling and robust fusion of multimodal temporal signals. To address the limitations of existing methods—insufficient exploitation of cross-modal synergies, sensitivity to noise, and constrained fine-grained alignment capabilities—this paper proposes a dual-stage cross-modal alignment framework. Stage 1 develops vision-text and audio-text contrastive learning networks based on a CLIP architecture, achieving preliminary feature-space alignment through modality-decoupled pre-training. Stage 2 introduces a temporal-aware dynamic fusion module integrating Temporal Convolutional Networks (TCN) and gated bidirectional LSTM to capture macro-evolution patterns of facial expressions and local dynamics of acoustic features, respectively. A novel quality-guided fusion strategy further enables differentiable weight allocation for modality compensation under occlusion and noise. Experiments on the Hume-Vidmimic2 dataset demonstrate superior performance with an average Pearson correlation coefficient of 0.51 across six emotion dimensions on the validate set. Remarkably, our method achieved 0.68 on the test set, securing runner-up in the EMI Challenge Track of the 8th ABAW (Affective Behavior Analysis in the Wild) Competition, offering a novel pathway for fine-grained emotion analysis in open environments.
\end{abstract}

%% file: sec/1_intro.tex
\section{Introduction}
\label{sec:intro}

Emotional mimicry stands as one of the core behaviors in human social interactions, where individuals unconsciously replicate others' facial expressions, vocal tones, or body movements to rapidly establish emotional resonance and strengthen social bonds. This psychological mechanism not only holds significant implications for understanding human social behavior but also provides natural inspiration for affective computing technologies in artificial intelligence. With the rapid development of applications such as virtual assistants, emotional robots, and mental health assessment, enabling machines to accurately perceive and quantify dynamic changes in human emotions has become a central challenge in human-computer interaction research\cite{ding2022letr,ding2023speed}. Among these, automatic estimation of Emotional Mimicry Intensity (EMI), capable of capturing the continuity and nuanced variations in emotional transmission, has gradually emerged as a cutting-edge direction in affective computing.

However, existing methods still face significant bottlenecks in addressing EMI estimation tasks. First, emotional mimicry is inherently a multimodal temporal process – subtle changes in visual expressions and dynamic fluctuations in vocal prosody often intertwine nonlinearly. Current research predominantly employs unimodal analysis (e.g., facial video only) or simplistic feature concatenation strategies, resulting in underutilized inter-modal synergy. For instance, baseline methods using ViT\cite{dosovitskiy2021imageworth16x16words} for visual information achieve merely 0.09 average Pearson correlation coefficient, indicating that unimodal visual models struggle to capture temporal evolution patterns of emotional intensity. Second, fine-grained alignment of emotional intensity requires cross-modal semantic understanding capabilities. Although Wav2Vec2\cite{baevskiWav2vec20Framework2020} audio features combined with linear layers attain a relatively higher performance of 0.24, their unimodal nature limits robustness in complex scenarios (e.g., strong noise, occlusion). Furthermore, existing fusion strategies fail to account for differential information contributions across modalities at varying emotional phases, leaving multimodal potential underdeveloped.

To address these challenges, this study proposes a multimodal fusion-based EMI estimation framework with three core contributions: First, we implement a dual-stage training paradigm. During the first stage, we construct image-text and audio-text dual-path alignment networks based on the CLIP architecture\cite{radford2021learningtransferablevisualmodels}, leveraging its contrastive learning mechanism to achieve primary alignment between text-visual and text-audio modalities. In the second stage, a modality mapping layer facilitates secondary alignment across image-text-audio triples. Compared to conventional approaches that separately extract features using three independent modality encoders, CLIP's alignment properties enable feature extraction that preserves cross-modal characteristics. Second, we design quality-aware dynamic fusion to adaptively reduce the contribution weights of impaired modalities. Third, temporal features are extracted through a Temporal Convolutional Network (TCN)\cite{bai2018empiricalevaluationgenericconvolutional} with segment-wise average pooling to suppress noisy frame interference and generate robust global representations, while audio modality employs bidirectional LSTM\cite{650093} to encode spectral-temporal dependencies.

This study is based on the EMI track of the 8th ABAW (Affective Behavior Analysis in-the-wild) Competition\cite{kollias2019deep,kollias2019face,kollias2019expression,kollias2020analysing,Kollias2025,kolliasadvancements,kollias2024distribution,kollias20246th,kollias2021affect,kollias2021analysing,kollias2021distribution,kollias2022abaw,kollias2023abaw,kollias2023abaw2,kollias2023multi,kollias20247th}. The adopted dataset, Hume-Vidmimic2, comprises 15,000 videos totaling over 25 hours of content.Participants in these videos mimic emotional expressions observed in seed videos, with their performance subsequently evaluated across multiple dimensions including "admiration," "amusement," "determination," "empathic pain," "excitement," and "joy."

Through systematic experiments on the Hume-Vidmimic2 dataset, this paper validates the effectiveness of the proposed method. Compared to baseline models, our framework achieves a breakthrough performance with an average Pearson correlation coefficient of 0.51 across six emotional dimensions, demonstrating 112\% improvement over the best baseline (0.24).

The remainder of this paper is organized as follows: Section 2 reviews related work. Section 3 details the employed multimodal features and model architecture. Section 4 describes experimental implementation specifics and presents result analyses. Finally, Section 5 concludes our work.

%% file: sec/2_related_work.tex
\section{Related Work}
\label{sec:relate}

\subsection{Multimodal Emotion Recognition Techniques}
Multimodal emotion recognition enhances the robustness of affective analysis by fusing complementary information from visual, audio, and textual modalities. Early work such as Dynamic Fusion Network (DFN) and Tensor Fusion Network (TFN)\cite{xu2018dynamicfusionnetworksmachine,zadeh2017tensorfusionnetworkmultimodal}, which dynamically weights features through inter-modal gating mechanisms, yet its static weight allocation struggles to adapt to temporal variations in emotional intensity. Recent studies focus on cross-modal interaction modeling: Multimodal Transformer (MulT)\cite{tsaiMultimodalTransformerUnaligned2019} attends to interactions between multimodal sequences across distinct time steps and latently adapt streams from one modality to another,but fails to explicitly address semantic discrepancies caused by modal heterogeneity. MISA~\cite{hazarikaMISAModalityInvariantSpecific2020} projects each modality into two distinct subspaces. The first subspace learns their commonalities and reduces inter-modal discrepancies, while the second subspace captures the modality-specific characteristics of each modality.It exhibits excellent generalization capabilities across diverse sentiment analysis tasks. Compared to these approaches, our framework combines TCN\cite{lea2016temporalconvolutionalnetworksunified} and bidirectional LSTM\cite{zhangBidirectionalLongShortTerm2015} to respectively model local-global temporal dependencies in visual and audio modalities, while employing segment-wise average pooling and modality-aware dynamic fusion mechanism to suppress interference from noisy frames. 

\subsection{Cross-Modal Alignment Techniques}
Cross-modal alignment aims to establish shared semantic spaces for multimodal data. CLIP\cite{radford2021learningtransferablevisualmodels} achieves image-text semantic alignment through contrastive learning, yet without extending to the audio modality. CLAP\cite{wuLARGESCALECONTRASTIVELANGUAGEAUDIO} constructed an audio-text joint embedding space based on the CLIP architecture, but their single-stage alignment strategy fails to balance multimodal semantic consistency. Recent work introduced the Flamingo model\cite{alayracFlamingoVisualLanguage2022}, achieving fine-grained image-text alignment through cross-attention, though its computational complexity limits practical deployment in real-time tasks. In affective computing, ImageBind\cite{girdharImageBindOneEmbedding2023a} learns a joint embedding across six different modalities - images, text, audio, depth, thermal, and IMU data. It enables novel emergent applications ‘out-of-the-box’ including cross-modal retrieval, composing modalities with arithmetic, cross-modal detection and generation.This paper proposes a dual-stage alignment framework: The first stage independently aligns image-text and audio-text pairs through CLIP and CLAP architectures, while the second stage introduces a modality mapping layer to optimize triple-modality joint semantics, significantly enhancing fine-grained consistency in emotional intensity estimation.

%% file: sec/3_method.tex
\section{Method}
The proposed Dynamic Multimodal Emotional Mimicry Intensity (EMI) Estimation Framework adopts a dual-stage training paradigm, integrating cross-modal alignment with hierarchical fusion mechanisms to address challenges in multimodal temporal modeling, dynamic weight allocation, and fine-grained semantic alignment. A detailed exposition follows from four technical perspectives: feature extraction, alignment strategies, fusion design, and training optimization.

\subsection{Multimodal Data Preprocessing}  
The input data comprises three modalities: visual, audio, and textual. The preprocessing methods for each modality are detailed below.

\textbf{Visual Modality}: Video streams are sampled at 5 frames per second, ensuring coverage of the typical dynamic range of human facial expressions (e.g., micro-expressions lasting approximately 1/3 second). Each frame is processed through the CLIP visual encoder (ViT-B/16) to extract 768-dimensional feature vectors, generating temporal feature sequences$V=\{v_1, v_2, ..., v_T\}$. ViT is preferred over CNN due to its global attention mechanism, which better captures coordinated semantic patterns in facial expressions (e.g., synchronized eyebrow relaxation and lip-corner raising associated with "admiration").

\begin{figure*}[htbp]
    \centering
    \includegraphics[width=\textwidth]{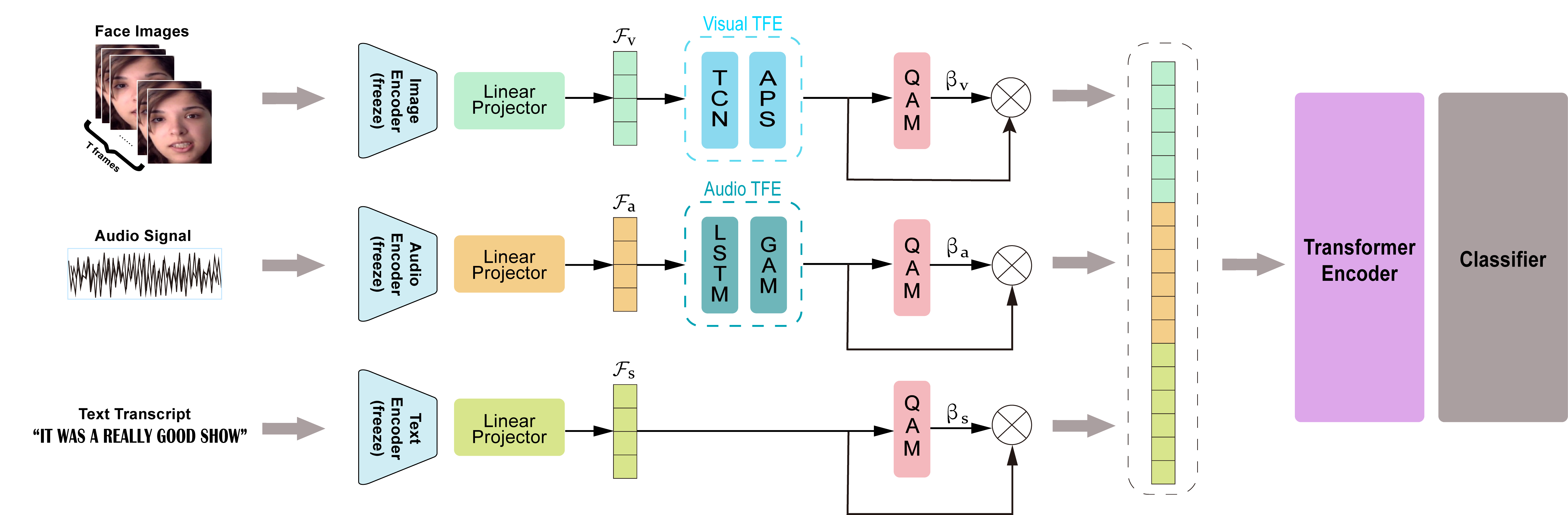}
    \caption{The overview of our proposed method. We first utilize the three encoders with frozen parameters pre-trained in the first stage, the audio, image, and text are then fed into these three encoders to obtain $F_A$, $F_I$, $F_T$. After that, $F_I$ and $F_A$ will be strengthened by temporal features for the Temporal Feature Enhancer (TFE), the temporally enhanced features are then processed by the Quality-Aware Module (QAM), which outputs weights for each feature. These weights are subsequently applied to the corresponding features, and the weighted features from the three modalities (audio, visual, text) are concatenated and fed into the Transformer Decoder module followed by the Classifier module to generate the final prediction.}
    \label{fig:pipeline}
\end{figure*}

\textbf{Audio Modality}: Raw audio signals are segmented into 20ms windows strictly aligned with video frames. Acoustic features are extracted using the pretrained Wav2Vec2.0 model, producing 1024-dimensional feature sequences $A=\{a_1, a_2, ..., a_T\}$. The model's self-supervised pretraining on unlabeled speech data enhances its representational capacity for emotion-related acoustic attributes like pitch and speaking rate.

\textbf{Text Modality}: The text processing pipeline begins with speech-to-text conversion using the Whisper model~\cite{radford2022robustspeechrecognitionlargescale}, which employs an end-to-end deep learning architecture optimized for speech recognition. Unlike conventional ASR systems, Whisper innovatively adopts a multi-task training paradigm, simultaneously optimizing speech recognition, language identification, and voice activity detection across 1.28 million hours of diverse audio data. This approach significantly improves robustness in noisy environments and multilingual adaptability. Specifically, we utilize a pruned and fine-tuned version, Whisper-large-v3-turbo, for high-accuracy transcription to obtain semantically complete utterances $s$.

Transcribed texts are encoded through the pretrained BERT model~\cite{devlin2019bertpretrainingdeepbidirectional} using its 12-layer Transformer architecture to extract 768-dimensional context-aware feature vectors $f_s$. As a foundational NLP model, BERT's dual pretraining objectives (masked language modeling and next-sentence prediction) enable effective capture of deep semantic relationships, providing reliable linguistic representations for our framework.

\subsection{Overall Framework}  
The proposed multimodal EMI estimation framework incorporates a dual-stage training paradigm and multimodal fusion architecture. The first stage achieves independent semantic alignment between image-text and audio-text pairs through contrastive learning. The second stage constructs cross-modal temporal modeling and dynamic fusion modules while freezing modality encoders. Final emotional mimicry intensity estimation is implemented end-to-end via a multi-layer Transformer encoder. Figure~\ref{fig:pipeline} shows an overview of the proposed framework.

\subsection{Dual-Stage Cross-Modal Alignment}  

\textbf{Stage I: Modality-Text Contrastive Alignment}
Building upon the CLIP architecture~\cite{radford2021learningtransferablevisualmodels}, we establish dual-path contrastive learning networks for vision-text and audio-text pairs. For image-text modality alignment, we utilize the Aff-Wild2\cite{kollias2019affwild2extendingaffwilddatabase} multimodal dataset. This database contains 548 in-the-wild videos (approximately 2.7 million frames) annotated with six basic expressions (anger, disgust, fear, happiness, sadness, surprise), the neutral state, and an "other" compound affective category. Additionally, each frame includes Action Unit (AU) annotations and Valence-Arousal (VA) annotations to capture facial muscle movements and emotional intensity dynamics. 

To address the underutilization of fine-grained affective semantics in existing facial expression recognition methods, we propose a cross-modal contrastive learning framework leveraging AffectNet's multimodal annotations.
We first integrate discrete emotion labels (8 basic emotions), continuous Valence-Arousal (VA) values, and Action Unit (AU) codes into hierarchical textual descriptions. For each facial image, we generate structured text prompts following this template:  

\begin{equation}
\begin{split}
\mathcal{T} &= \underbrace{\text{"[Expression Class]"}}_{\text{Primary Label}} \\
           &\quad + \underbrace{\text{"with [AU Codes]"}}_{\text{Muscular Activity}} \\
           &\quad + \underbrace{\text{"Valence=VA1, Arousal=VA2"}}_{\text{Psychological Dimension}}
\end{split}
\end{equation}
For instance: "Happy (Intensity: High) with AU6 (Cheek Raiser) and AU12 (Lip Corner Puller), Valence=0.83, Arousal=0.65". This multi-granularity text augmentation converts the original sparse annotations into semantically dense text.

Architecturally, we adapt the CLIP framework for cross-modal alignment: 
A VIT-large visual encoder extracts facial features $f_v \in \mathbb{R}^{768}$, while a BERT-base text encoder generates textual embeddings $f_s \in \mathbb{R}^{768}$. 
The contrastive loss incorporates annotation confidence weights into InfoNCE:  
\begin{equation}
\mathcal{L}_{\text{con}} = -\frac{1}{N}\sum_{i=1}^N w_i \log\frac{\exp(f_v^{(i)} \cdot f_s^{(i)}/\tau)}{\sum_{j=1}^N \exp(f_v^{(i)} \cdot f_s^{(j)}/\tau)}
\end{equation}
where weights $w_i$ are determined by VA value standard deviations, with $\tau=0.07$ as temperature. 

For audio-text modality alignment, we use both Aff-Wild2 and Hume-Vidmimic2\cite{Kollias2025} dataset.
Specially, for audio modality, the CLAP architecture~\cite{wuLARGESCALECONTRASTIVELANGUAGEAUDIO} processes acoustic signals through Wav2Vec2.0~\cite{baevskiWav2vec20Framework2020} ,which extract audio features $f_a \in \mathbb{R}^{d_a}$. The remaining parts follow the same image-text alignment process as described above.

\noindent\textbf{Stage II: Tri-Modal Joint Optimization}
With frozen first-stage vision/audio/text encoders, we construct cross-modal mapping layers. Video features $\{f_v^t\}$ are processed by Temporal Convolutional Networks (TCN) to capture multi-scale temporal dependencies through stacked dilated convolutions:  
\begin{equation}  
h_v^t = \text{TCN}(f_v^t; \{\mathbf{W}_k^{d_k}\}_{k=1}^K)  
\end{equation}  
where $d_k$ indicates the dilation rate of the $k$-th convolutional layer and $K$ the total layer count. Audio features $\{f_a^t\}$ are fed into bidirectional LSTM to model spectral-temporal dynamics:  
\begin{equation}  
h_a^t = \text{BiLSTM}(f_a^t; \mathbf{W}_f, \mathbf{W}_b)  
\end{equation}  
Through modality mapping layers (MLP), $h_v^t$, $h_a^t$ and text features $f_s$ are projected into a shared space and concatenated along temporal dimension:  
\begin{equation}  
H_{\text{fuse}}^t = [\text{MLP}(h_v^t); \text{MLP}(h_a^t); \text{MLP}(f_s)]  
\end{equation}  

\subsection{Temporal Feature Enhancement}  
\textbf{Visual Modality Processing:} Building upon TCN outputs $h_v^t$, we employ a segment-wise average pooling strategy: partition videos into $M$ equal-length segments $\{C_m\}_{m=1}^M$, then compute per-segment feature means:  
\begin{equation}  
\hat{h}_v^m = \frac{1}{|C_m|} \sum_{t \in C_m} h_v^t  
\end{equation}  
This operation suppresses local noise while preserving critical frame information for emotional intensity evolution.  

\noindent\textbf{Audio Modality Processing:} A gated attention mechanism is applied to BiLSTM outputs $h_a^t$:  
\begin{equation}  
\alpha_a^t = \sigma(\mathbf{W}_g [h_a^t; \Delta h_a^t])  
\end{equation}  
where $\Delta h_a^t = h_a^t - h_a^{t-1}$ denotes adjacent-frame feature differences, and $\sigma$ represents the sigmoid function. The enhanced features are calculated as $\tilde{h}_a^t = \alpha_a^t \odot h_a^t$.

\subsection{Quality-Aware Dynamic Fusion}  
Recent study~\cite{hallmenUnimodalMultiTaskFusion2024} conducted a systematic multimodal robustness analysis of the Hume-Vidmimic2 dataset. By constructing a hierarchical evaluation framework under visual quality degradation scenarios, they revealed the significant impact of Facial Visual Quality (FVQ) on multimodal affective computing models. Critical interference factors such as partial occlusions (e.g. hair occluding eye/eyebrow regions) and abnormal illumination fluctuations (side overexposure-induced facial highlight noise) were shown to increase error rates in facial Action Unit (AU) detection and further induce cross-modal representation misalignment phenomena.

To mitigate the impact of inevitable noise in individual modalities, we design a modality quality assessment module for dynamic adjustment of multimodal contribution weights. For timestep $t$, modality quality scores are computed as:  
\begin{equation}  
\ q_m = \text{MLP}(h_m), \ m \in \{v,a,s\} \\  
\end{equation}  
Dynamic fusion weights are generated via softmax normalization:  
\begin{equation}  
[\beta_v^t, \beta_a^t, \beta_s^t] = \text{softmax}([q_v^t, q_a^t, q_s])  
\end{equation}  
The weighted fused features are fed into a multi-layer Transformer encoder for emotional intensity prediction:  
\begin{equation}  
\hat{y}^t = \text{Transformer}(\beta_v^t \hat{h}_v^m + \beta_a^t \tilde{h}_a^t + \beta_s^t f_s)  
\end{equation}  
This design enables adaptive attenuation of impaired modalities' contributions when quality discrepancies exist in multimodal signals (e.g., facial occlusion or environmental noise).

\subsection{Training Strategy}  
We adopt a two-stage training paradigm: Stage I optimizes modality-text alignment using contrastive loss $\mathcal{L}_{v2t}+\mathcal{L}_{a2t}$; Stage II freezes encoder parameters and optimizes regression tasks through mean squared error loss $\mathcal{L}_{\text{MSE}} = \frac{1}{T}\sum_{t=1}^T (y^t - \hat{y}^t)^2$.

%% file: sec/4_experiment.tex
\section{Experiment}
\subsection{Datasets and Experimental Setup}  
Experiments are conducted on the Hume-Vidmimic2\cite{Kollias2025} multimodal affective dataset, which contains 15,000 video samples spanning six emotional dimensions ("Admiration", "Amusement", "Determination", "Empathic Pain", "Excitement", "Joy"). Each sample's emotional intensity score, ranging within $[0,1]$, is annotated by professional annotation teams following standardized protocols. The dataset is partitioned into training, validation, and test sets. Statistical details of each split are provided in Table ~\ref{tab:dataset}.
\begin{table}[htbp]
  \centering
    \begin{tabular}{ccc}
    \toprule
    Partition & Duration & Samples \\
    \midrule
    Train & 15:07:03 & 8072 \\
    Validation & 9:12:02 & 4588 \\
    Test  & 9:04:05 & 4586 \\
    \midrule
       $\sum$   & 33:23:10 & 17246 \\
    \bottomrule
    \end{tabular}%
    \caption{Hume-Vidmimic2 partition statistics.}
  \label{tab:dataset}%
\end{table}%

\subsection{Implementation Details}  
\textbf{Evaluation Metrics} The mean Pearson correlation coefficient (\(\rho\)) serves as the primary evaluation metric for intensity estimation, measuring the linear correlation between predicted emotional response intensities and ground truth values. This metric is defined as:  
\[
\rho = \sum_{i=1}^{6} \frac{\rho_{i}}{6} 
\]  
where \(\rho_{i} \ (i \in \{1,2,\cdots,6\})\) denotes the correlation coefficient for each of the six emotion categories, calculated as:  
\[
\rho_{i} = \frac{\text{cov}(y_{i}, \hat{y_{i}})}{\sqrt{\text{var}(y_{i}) \text{var}(\hat{y_{i}})}} \tag{9}
\]  
Here: \(\text{cov}(y_{i}, \hat{y_{i}})\) represents the covariance between predicted and target values.\(\text{var}(y_{i})\) and \(\text{var}(\hat{y_{i}})\) denote the variances of target and predicted values respectively.

\subsection{Training Strategy Optimization}  
\textbf{Cosine Annealing Scheduling}:  
During the contrastive pretraining phase (50 epochs), we implement Stochastic Gradient Descent with Warm Restarts (SGDR) with learning rate updates governed by:  
\begin{equation}  
\eta_t = \eta_{\min} + \frac{1}{2}(\eta_{\max} - \eta_{\min})\left(1 + \cos\left(\frac{T_{\text{cur}}}{T_i}\pi\right)\right)  
\end{equation}  
where $\eta_{\max}=1\times10^{-5}$, $\eta_{\min}=1\times10^{-7}$, $T_i$ denotes cycle length (10 epochs), and $T_{\text{cur}}$ counts iterations within current cycle. Each cycle concludes with learning rate resetting while preserving optimizer momentum via warm restarts to accelerate convergence.

\noindent\textbf{EMA Exponential Smoothing}:  
Parameter-level Exponential Moving Average (EMA) is activated during fine-tuning with momentum decay factor $\gamma=0.999$. The update rule follows:  
\begin{equation}  
\theta_{\text{EMA}}^{(t)} = \gamma \theta_{\text{EMA}}^{(t-1)} + (1-\gamma)\theta^{(t)}  
\end{equation}  
where $\theta^{(t)}$ represents model parameters at step $t$, and $\theta_{\text{EMA}}^{(t)}$ denotes smoothed parameters. Validation-phase inference employs EMA parameters to suppress training oscillations and enhance test-set generalization.

\section{Results}  

\subsection{Unimodal Performance Analysis}
As shown in Table~\ref{tab:unimodal}, our enhanced unimodal models demonstrate consistent improvements over baseline implementations across both modalities. For the visual modality, it can be observed that our ViT model pre-trained using contrastive learning demonstrates significantly superior performance compared to the official ViT features. our modified ViT architecture achieves a Pearson correlation coefficient ($\rho_{val}$) of 0.1481, representing a 70\% relative improvement over the baseline's 0.0873. In the acoustic domain, our optimized Wav2Vec2.0 implementation reaches $\rho_{val}$=0.2734, outperforming the baseline (0.2405) by 13.9\%. This significant performance gain in audio processing suggests our feature enhancement strategy effectively captures subtle paralinguistic cues critical for emotional mimicry estimation.

\begin{table}[htbp]
  \centering
    \begin{tabular}{ccc}
    \toprule
    Features  & Modality & Mean $\rho_{val}$ \\
    \midrule
    ViT(baseline)\cite{kollias20246th} & $V$     & 0.0873 \\
    ViT(ours) & $V$     & \textbf{0.1481} \\
    \midrule
    Wav2Vec2.0(baseline)\cite{kollias20246th} & $A$     & 0.24 \\
    Wav2Vec2.0(ours) & $A$     & \textbf{0.2734} \\
    \bottomrule
    \end{tabular}%
    \caption{The unimodal results on validation set of the EMI Estimation Challenge. We report the Pearson correlation coefficient ($\rho$) for the average of 6 emotion targets. Where $V$ represents the visual modality, and $A$ represents the acoustic modality.}
  \label{tab:unimodal}%
\end{table}%

\subsection{Multimodal Fusion Effectiveness}
As shown in Table~\ref{tab:ablation}, the Quality-Aware Module (QAM) demonstrates critical efficacy in multimodal emotion analysis. The baseline model (VIT+Wav2Vec+BERT) achieves a Pearson correlation coefficient $\rho_{val}$ of 35.05\%, while introducing QAM alone boosts performance to 48.81\% (+39.3\% relative improvement).This significant performance improvement stems from QAM's selective feature activation mechanism. When the video modality is corrupted by noise, QAM dynamically shifts fusion weights toward more stable audio and text modalities.

\begin{table}[htbp]
  \centering
  \begin{tabular}{lcccc}
    \toprule
    Experimental Combination & TFE & QAM & Mean $\rho_{val}$  \\
    \midrule
    VIT+Wav2Vec+BERT & & &  0.3505 \\
    +TFE & \checkmark &  &  0.4219 \\
    +QAM &  & \checkmark &  0.4881 \\
    +TFE + QAM & \checkmark & \checkmark &  \textbf{0.5126} \\
    \bottomrule
  \end{tabular}
  \caption{Module Ablation Experiment Results on $\rho_{val}$.The 'VIT+Wav2Vec+BERT' combination serves as our baseline.}
  \label{tab:ablation}
\end{table}

\subsection{Multimodal Fusion Effectiveness}
Table~\ref{tab:multimodal} reveals the progressive performance enhancement through multimodal integration. Our key findings include:

\begin{itemize}
\item The audio-visual pair achieves $\rho_{val}$=0.4846, surpassing visual-textual fusion (0.2834) a lot.This indicates that audio data contributes disproportionately more than textual data.
\item The audio-textual pair demonstrates strong synergy with $\rho_{val}$=0.4918, highlighting text modality's complementary role
\item The full multimodal integration achieves the best performance ($\rho_{val}$=0.5126),
\end{itemize}

\begin{table}[htbp]
  \centering
    \begin{tabular}{cccccc}
    \toprule
    Visual & Audio & Text  & Mean $\rho_{val}$ \\
    \midrule
     ViT & - & - & 0.2683 \\
     ViT & - & BERT & 0.2834 \\
     - & Wav2Vec2.0 &   -    & 0.4703 \\
     ViT & Wav2Vec2.0 &   -    & 0.4846 \\
     - & Wav2Vec2.0 & BERT & 0.4918 \\
     ViT & Wav2Vec2.0 & BERT & \textbf{0.5126} \\
    \bottomrule
    \end{tabular}%
    \caption{The Multimodal results on validation set of the EMI Estimation Challenge.We use our pre-trained encoders for three modalities with both TFE and QAM to conduct ablation experiments.}
  \label{tab:multimodal}%
\end{table}%
The progressive performance gains confirm our hypothesis about cross-modal complementarity. Notably, the textual and audio modality contribute disproportionately to system performance, suggesting linguistic and audio context play a crucial role in disambiguating emotional mimicry patterns. Our dynamic fusion mechanism successfully preserves modality-specific information while enabling effective cross-modal interaction, as evidenced by the super-additive performance improvement in three-modal fusion.

%% file: sec/5_conclusion.tex
\section{Conclusion}

\label{sec:conclusion}
This paper presents our solution to the Emotional Mimicry Intensity (EMI) estimation challenge in the 8th Affective Behavior Analysis in-the-wild (ABAW) Competition. By introducing a dual-stage training paradigm and quality-aware dynamic modality fusion mechanism, our method achieves enhanced robustness in EMI estimation while maintaining tight three-modality alignment. Systematic validation on the Hume-Vidmimic2 dataset demonstrates that our approach attains an average Pearson correlation coefficient of 0.51 across six emotional dimensions.Moreover, our method achieved an average Pearson correlation coefficient of 0.68 on the test set, secured the runner-up position in the 8th ABAW (Affective Behavior Analysis in the Wild) Competition, demonstrating excellent performance in emotion recognition tasks under uncontrolled real-world conditions. Provide a new approach for emotional mimicry intensity estimation.